\pdfoutput=1

\documentclass[11pt]{article}

\usepackage{ACL2023}

\usepackage{graphicx}
\usepackage{subcaption}
\usepackage{times}
\usepackage{latexsym}
\usepackage{float}
\usepackage{placeins}
\usepackage{xcolor}
\usepackage[T1]{fontenc}

\usepackage[utf8]{inputenc}

\usepackage{microtype}

\usepackage{inconsolata}
\usepackage{comment}
\usepackage{siunitx}
\title{Building a Corpus for Biomedical Relation Extraction of Species Mentions} 

 \author{Oumaima El Khettari, Solen Quiniou, Samuel Chaffron \\
         Nantes Université, École Centrale Nantes, CNRS, LS2N, UMR 6004, F-44000\\
         \{\texttt{oumaima.el-khettari, solen.quiniou, samuel.chaffron\}@ls2n.fr} \\
    }

\begin{document}

\maketitle

\begin{abstract}
We present a manually annotated corpus, Species-Species Interaction, for extracting meaningful binary relations between species, in biomedical texts, at sentence level, with a focus on the gut microbiota. The corpus leverages PubTator to annotate species in full-text articles after evaluating different Named Entity Recognition species taggers. Our first results are promising for extracting relations between species using BERT and its biomedical variants.
\end{abstract}

\section{Introduction}

The field of biomedical relation extraction (RE) has made significant advancements in recent years, with the development of various state-of-the-art models for extracting meaningful relationships between entities from scientific articles. However, the availability of annotated datasets for specific types of relations, such as interactions between species, remains limited. Studying the interactions and relationships between different species is of great interest in many biomedical research areas, including gut microbiome research. 
The gut microbiota, a complex community of microorganisms residing in the human gastrointestinal tract, plays a crucial role in human health and disease. Understanding the interactions and relationships between different species within the gut microbiome is of significant importance in advancing knowledge on microbiome-mediated health outcomes. 

The first challenge in studying this type of interactions is the lack of annotated datasets that specifically capture the semantic expressions of interactions between species. Most studied interactions occur between proteins, drugs, genes, and diseases. The representations of these relations may not fully capture the complexity of species interactions, particularly in the specific context of the microbiome. 
In this article, we present a dedicated corpus for relation extraction of species mentions in biomedical texts, by manually annotating binary relations at the sentence level in an existing corpus for Named Entity Recognition (NER) and in a new corpus focusing on the gut microbiome. We also choose to leverage PubTator, a reliable system for annotating species in full-text articles, after evaluating its performance with respect to a fine-tuned BERT and a scispaCy model. We carefully selected a set of PMC documents from PubTator based on heuristics that align with our domain of interest and annotated a selection of sentences from the corpus with binary relations, consisting of the presence/absence of a relation. Afterwards, we proceeded to fine-tune existing transformer-based models on our corpus to highlight the impact of a new small set of semantic relation expressions. Our contributions are as follows:

\begin{itemize}
    \item A study of the Species entities in the literature;
    \item Species-Species Interaction (SSI), a corpus of manually annotated binary relations on full-text scientific articles, publicly available on HuggingFace\footnote{\href{https://huggingface.co/datasets/taln-ls2n/SSI}{https://huggingface.co/datasets/taln-ls2n/SSI}};
    \item A comparative study of the SOTA Species NER taggers;
    \item Initial results on extracting relations between species using BERT and several of its biomedical variants.
\end{itemize}

\section{Related Work}

\subsection{NER for Species}
The presence of relations in a text directly implies the presence of named entities of interest. Our work is based on the hypothesis stating the following: a relation cannot be expressed in a text unless at least two named entities are mentioned in the same text. Prior work involving the investigation of species mentions has been conducted in order to evaluate existing methods of NER, for the entity type "Species".

Little research has been conducted on the specific Species mention in the context of NER and RE. This scarcity can be attributed primarily to the lack of resources specifically designed for Species. First, methods applied for the mention Species were rule-based systems along with dictionary-based models, together with statistical methods, such as TaxonFinder~\cite{taxonfinder} and MetaMap~\cite{metamap}. This type of method strongly depends on the completeness of ontologies, knowledge graphs, and dictionaries. Then, deep learning methods requiring massive amounts of annotated data, relied exclusively on a few available corpora of species mentions to fine-tune several biomedical variants of BERT~\cite{devlin2018bert}, such as BioBERT~\cite{BioBERT}. 

In this context, scispaCy~\cite{ScispaCy}, a library specifically designed for biomedical and scientific text mining, offers a variety of NER models, built on the deep learning spaCy models. PubTator Central~\cite{pubtator}, a web-based tool designed to automatically annotate and extract information from biomedical literature, provides access to more than 3 million full-text articles that have been annotated with different entities, including Species~\cite{pubtator2}. Its annotations are generated using a combination of dictionary- and machine learning-based models trained on large biomedical data. 

In terms of corpora for NER, LINNAEUS~\cite{linnaeus} and S800~\cite{species} are the most widely known and used corpora for Species mentions as they served for the fine-tuning of deep learning models. LINNAEUS consists of 100 randomly selected and manually annotated full-text documents from the PMC~OA document set, resulting in 2\,988 annotations. As for S800, it is a collection of 800 randomly collected and manually annotated abstracts from journals, classified into 8 categories, and including 3\,708 annotations.

\subsection{Relation Extraction in Biomedical Literature}
Relation Extraction is a task where the goal is to identify and classify the relationships between pairs of entities. This specific task is modeled as a classification problem. In binary relation extraction, the task is to determine whether a relationship exists between two entities in a given sentence. The goal is to classify the sentence as either containing a relationship or not. Another task is to identify the type of relationship between two entities, among some predefined categories. For each type of relation extraction, a suitable corpus has to be defined.

There are several existing biomedical RE corpora that have been widely used for the training and fine-tuning of models, where a variety of interaction types are captured. For most of these corpora, the provided annotations are binary ones, indicating the presence or absence of a relation, at the sentence level, for protein-protein interactions (PPI)~\cite{hamilton2017inductive}, drug-drug interactions (DDI)~\cite{DDI}, gene-disease interactions (GAD)~\cite{Bravo2015}, as well as EU-ADR~\cite{euadr}.

Despite the large diversity of the above entities in existing datasets, the mention Species remains off the list. Consequently, interactions between species remain under-represented, and the diverse nature of species-related information, specifically in scientific articles, including taxonomy differences, evolutionary relationships, and ecological interactions, further complicates the task of relation extraction in this domain.

Biomedical RE dataset labeling is known to be challenging, often time-consuming, and requiring domain expertise. As a result, manually labeled datasets are often size limited. Due to these challenges, distant supervision has been introduced as a strategy for data augmentation in biomedical RE~\cite{mintz2009distant}. While this strategy proved its efficiency for some use cases, it hinders the extraction of semantic relations.
As for NER, a variety of methods can perform RE, starting from rule-based models to transformer-based ones. BERT and its biomedical variations are achieving promising results for this task, especially on the binary relation extraction task, using the corpora previously cited.

\section{Relation Extraction Corpus for Species}

The participating entities in known interactions are major biomedical concepts, such as genes, proteins, diseases, and drugs. As we focus on studying relations occurring between species in scientific research articles, we encountered significant challenges. One major challenge is the scarcity of annotated corpora that capture the nuanced semantic expressions of interactions between species, which are distinct from the well-studied RE datasets. As a first step to tackle this problem, we introduced Species-Species Interaction (SSI), a manually annotated corpus dedicated to species binary interactions, built from full-text scientific articles.

\subsection{Selecting Initial Corpora}
Our main goal is to extract information from full-text articles, as we aim to detect interactions that are poorly represented and may not be found in smaller parts of texts, like abstracts. 
We were interested, at the same time, by extending existing corpora mentioning Species (aka S800 and LINNAEUS) by adding Species-Species relations as well as creating a new corpus specifically dedicated to the gut microbiome, using PubTator to provide Species mentions. 

The selection of PubTator articles was thus based on the following query: \textsc{<"gut microbiota" OR "gut microbiome" OR "intestinal microbiota" OR "intestinal microbiome">}. Then, to select relevant articles for our domain, we established heuristics taking into account the number of species mentions in the NCBI taxonomy~\cite{ncbi} and the diversity of these annotations. We also considered the hierarchical structure of the taxonomy, as common non-specialized vocabulary terms like "human," "patient," or "mice" are frequently used in scientific articles. As these terms most often refer to model organisms, they may not be relevant to our study.

\subsection{Annotating Species-Species Relation}
The step of collecting data is followed by splitting the articles into sentences using spaCy, since the annotation is at the sentence level. We focused on sentences where exactly two species are mentioned to maximize the probability that a binary relation is indeed expressed. Consequently, no sentences with the previous criterion were found on S800. This is due to the sparsity of entities mentions in abstracts. Therefore, we collected 442 sentences from the LINNAEUS corpus and 557 sentences from the corpus obtained with PubTator, to obtain our SSI corpus. 
\paragraph{Annotation Guidelines}
We aspired to capture the semantic representation of interactions between species, in sentences. 
We refer to interactions between two species as various mechanisms through which the species impact one another's behaviour, survival, or the overall ecological dynamics within an ecosystem.

Before the annotation process, we applied the mask @SPECIES\$ on the species mentions in all sentences. The objective here was to limit the impact of external knowledge, like the name of a species, on the decision of the presence of a relation. 
We finally assigned the label 1 to sentences containing a relation between at least two species mentions (to denote the presence of a relation) and the label 0 when no relation seems to appear in the sentences.

\begin{table}[hbt!]
    \centering
    \begin{tabular}{|p{5cm}|p{1cm}|}
        \hline \textbf{Sentences} &\textbf{Label} \\
        \hline
         \emph{Effects of dark @SPECIES\$ and @SPECIES\$ consumption on endothelial function and arterial stiffness in overweight adults} & 0 \\
         \hline
         \emph{One @SPECIES\$ (Epi132) revealed family cancer occurrence resembling families harboring CHEK2 mutations in general, the other @SPECIES\$ (epi203) was non-conclusive.} & 0 \\
        \hline
         \emph{Eosinophil accumulation induced by @SPECIES\$ interleukin-8 in the @SPECIES\$ in vivo.} & 1\\
         \hline
         \emph{@SPECIES\$ ellagitannins thwarted the positive effects of dietary fructo-oligosaccharides in @SPECIES\$ cecum.} & 1\\
    \hline
    \end{tabular}
    \caption{SSI annotation examples}
    \label{annotation}
\end{table}

In table~\ref{annotation}, we give examples of annotated sentences.
A set of annotation guidelines have been defined so as to achieve the previously mentioned goal while annotating, including the following:
\begin{itemize}
    \item Contextual indicators: The first step of the annotation process is analyzing the sentence for any contextual cues suggesting a relation between the species. This can include explicit keywords, verbs, or phrases that imply a connection, such as "interacts with", "physically related to", "association".
    \item Negation and absence: Negation is considered as a type of relation. Thereby, sentences with a clear identifiable negation should be annotated with label 1.
    \item Comparison: Sentences claiming a comparison between the mentioned species should be labeled as 1 only if they share a clearly cited characteristic.
    \item Multiple relations: Only consider the potential presence or absence of a relation between the masked species. 
\end{itemize}

Table~\ref{tableSSI} gives the statistics on the train, validation, and test splits of our SSI corpus. We used 60\% of the corpus for the train set and we equally split the remaining 40\% on the test and validation sets.

\begin{table}[h!tb]
\centering
\begin{tabular}{llll}
\hline
\textbf{SSI} & \textbf{\#train} & \textbf{\#val} & \textbf{\#test} \\
\hline
label 1 & 300 & 93 & 109  \\
label 0 & 299 & 107 & 91  \\
\hline
\end{tabular}
\caption{Statistics on our corpus where label 0 states for the absence of a relation and label 1 for its presence}
\label{tableSSI}
\end{table}

\section{Experiments}
We first conducted experiments on NER taggers for the Species mentions and then evaluated BERT-like models for the RE tasks on the SSI corpus.

\subsection{Evaluating NER Taggers on Species}
As a preliminary study of the Species RE and in order to add NER annotations to our RE corpus, we compared the performance of fine-tuned BERT on LINNAEUS, scispaCy model, and PubtaTor on S800 for the task of NER. One of NER scispaCy models is en\_ner\_craft\_md. It is trained on the CRAFT corpus~\cite{bada2012concept} and it identifies various biomedical entities, including taxons, a close term to species. A taxon refers to a group of organisms classified based on shared characteristics, while species is the basic unit of biological classification consisting of individuals that can interbreed and produce fertile offspring. Due to the proximity of these two concepts, we consider taxon a reference to species.

In addition to precision, recall, and F1-score, we report the Slot Error Rate (SER). It is a metric that complements the previous measurements by considering various factors, including deletions, insertions, and boundary-related errors.

During the comparison of the collected named entities, an issue was encountered while matching S800 articles with their versions from PubTator due to encoding problems. As a result, the text was garbled or displayed as gibberish characters, with certain letter combinations replaced with symbols. This was particularly problematic when the affected text was a named entity. To address this issue, spaces were corrected and the texts were categorized into two classes: the first containing the majority of articles with matching versions, and the second containing articles with encoding problems. The evaluation was performed on the first category, containing 2924 entity.

\begin{table}[t!b]
\centering
\begin{tabular}{lllll}
\hline
\textbf{Models} & \textbf{P} & \textbf{R} & \textbf{F1} & \textbf{SER}\\
\hline
Fine-Tuned BERT & 0.53 & 0.62 & 0.57 & 0.75 \\
ScispaCy & 0.22 & 0.22 & 0.22 & 1.35 \\
PubTator & 0.68 & 0.74 & 0.71 & 0.51 \\
\hline
\end{tabular}
\caption{Comparing NER models on Species mentions}
\label{tableNER}
\end{table}

As shown in Table~\ref{tableNER}, PubTator outperforms fine-tuned BERT on Linnaeus as well as the scispaCy model. It should be mentioned that the metrics are calculated on the exact matches of the named entities. However, the SER value, reaching a minimal value of 51\%, sheds light on the complexity of handling composed or long species mentions. Nevertheless, PubtaTor handles better these instances compared to the remaining systems. For this reason, we decided to use PubtaTor for the annotation of Species in our corpus.

\subsection{Comparing models for RE on Species}
We performed RE by fine-tuning existing models on the SSI corpus. Along with BERT-base, we used the following models, chosen for their promising results on other biomedical RE corpora: 
\begin{itemize}
    \item \textbf{BioBERT}, pre-trained on abstracts from the PubMed database and full-text articles from the PubMed Central Open Access Subset;
    \item \textbf{SciBERT}~\cite{beltagy2019scibert}, pre-trained on Semantic Scholar full-text articles, with a majority belonging to the biomedical domain;
    \item \textbf{BioLinkBERT}~\cite{yasunaga2022linkbert}, pre-trained on PubMed abstracts as well as their citations in order to enrich the model with the existing dependencies between the academic articles.
\end{itemize}

We report, in Table~\ref{tableRE}, the results on the test set of SSI, before and after the fine-tuning of the compared models. 

Prior to the fine-tuning on SSI, one major observation was that all models performed poorly on label 0, except BioLinkBERT. This is due to the fact that there are larger possibilities of semantic expressions for no relation. This is not the case for the second category as the set of relations between species is finite.  
On the contrary, BiolinkBERT achieved a very low recall on label 1, a tendency that was altered after the fine-tuning by gaining points on the F1 score.
Overall, the results indicated that fine-tuning pre-trained models led to a notable improvement in performance, as the precision and the F1 scores were higher for all the fined-tuned models as compared to their pre-trained counterparts.

\begin{table}[h!bt]
    \centering
    \begin{tabular}{|l|c|c|c|c|} 
        \hline
         Models & \multicolumn{2}{|c|}{Label 0} & \multicolumn{2}{|c|}{Label 1}\\
         & P & F1 & P & F1 \\
        \hline
        BERT base & 0.33 & 0.02 & 0.54 & 0.69 \\
        BioBERT & 0.39 & 0.26 & 0.55 & 0.62 \\
        SciBERT & 0.37 & 0.34 & 0.49 & 0.52 \\
        BioLinkBERT & 0.45 & 0.62 & 0.50 & 0.02 \\
        \hline 
        FT BERT base & 0.72 & 0.71 & 0.76 & 0.76 \\
        FT BioBERT & 0.64 & 0.68 & 0.75 & 0.70 \\
        FT SciBERT & 0.69 & 0.71 & 0.76 & 0.75  \\
        FT BioLinkBERT & 0.66 & 0.70 & 0.77 & 0.72 \\ 
        \hline
    \end{tabular}
    \caption{Results of RE before and after fine-tuning models on SSI (fine-tuned models are denoted FT)}
    \label{tableRE}
\end{table}

\subsection{Error Analysis}

Given the multiple demonstrations of the effectiveness of fine-tuning in enhancing downstream tasks, we conducted an error analysis on the test sample for each model. By this mean, we aimed to gain insights into the performance of the models, as well as the processing of the new corpus.

Upon analyzing the errors made by all the models, it became apparent that the majority of confusion arised when dealing with enumerated species mentions. In these cases, our annotation required the presence of a relation only if the enumerated species share a distinctly mentioned characteristic. However, accurately identifying and highlighting such characteristics poses a challenge. Errors also occurred in instances where the terms "relationship," "interaction," or negation indicators are explicitly stated in the text. Despite their explicit mention, the semantic information may not indicate a genuine interaction between the two species or clearly expresses a negation. 
Another significant concern pertains to co-references when examining examples with co-referent species mentions. According to the annotation, these instances are marked as having a relation. However, models tend to classify them as 0 due to the semantic challenge of identifying indicators that signify a relation.
Furthermore, our findings indicate that there is an increased likelihood of mislabeling, particularly in long sentences, as the distance between two species mentions grows in terms of the number of words.
Additionally, there is an error type closely associated with named entities. When the masked species mention is part of a compound name, it poses a challenge for the models to determine whether the relation is solely between the masked entity or the entire named entity.

Addressing these challenges is essential to enhance the performance of the models. Major indicators of improvement in our case include focusing on co-reference resolution, expanding the dataset by incorporating enumeration instances, varying the length of sentences, and further enhancing named entity recognition systems for the mention Species.

\section{Conclusion}
 
 We introduced a new corpus called Species-Species Interaction (SSI) for extracting meaningful binary relations between species in biomedical texts, with a focus on the gut microbiome, to address the absence of Species-related corpora. We included a comparative study of species NER taggers followed by initial results on extracting relations using BERT and its biomedical variants. Our experiments demonstrated that performances of pre-trained models were improved with fine-tuning, despite the use of a limited-size corpus.
To gain a better understanding of these results, we performed an error analysis, allowing us to identify specific areas for improvement that should be taken into consideration.
 In the future, we will aim to enlarge our SSI corpus by further annotating sentences and also by integrating relation types to better represent Species-Species interactions.

\section{Limitations}

The methodology of the constructed corpus is based on NER, making SSI a corpus integrating RE with NER. We aimed to enrich our experiments with multi-task learning so as to measure the impact of NER on the extraction of relations. Unfortunately, after conducting statistical analysis on SSI, it showed that, due to its limited size, we did not have enough instances of the same pair of entities participating in one sentence. As this statement would negatively bias our multi-task model, we refrained from conducting these experiments, before proceeding to annotate further instances and add them to our corpus in further work.
In terms of annotation, only one annotator has been invested in the current task, as it is a preliminary work including a new dataset for relation extraction, with new entities and relation types. 
In a future work, multiple annotators performing the annotation task and metrics for the corpus quality will be presented.

\section*{Acknowledgements}
We would like to express our gratitude to Gesser Riahi and Adrien Lesenechal for their valuable contributions to this work. Their commitment was instrumental in the successful completion of this project. 
We would also like to thank the anonymous reviewers for their valuable inputs to this article. 

This work was financially supported by the ANR AIBy4 project (ANR-20-THIA-0011) and Nantes Université.

\bibliography{acl2023}
\bibliographystyle{acl_natbib}

\appendix

\section{Training settings}

\label{Training settings}

\begin{itemize}
    \item GPU type: TITAN X 12Go
    \item Text max size: 512
    \item Learning rate: \num{1e-5}
    \item Batch size: 8
    \item Number of epochs: 10
    \item Optimizer: AdamW
\end{itemize}

\end{document}